\documentclass[conference]{IEEEtran}
\IEEEoverridecommandlockouts
\usepackage{cite}
\usepackage{amsmath,amssymb,amsfonts}
\usepackage{algorithmic}
\usepackage{graphicx}
\usepackage{textcomp}
\usepackage{xcolor}
\usepackage{cleveref}
\usepackage{subcaption}

\def\BibTeX{{\rm B\kern-.05em{\sc i\kern-.025em b}\kern-.08em
    T\kern-.1667em\lower.7ex\hbox{E}\kern-.125emX}}
\begin{document}

\title{Speed-based Filtration and DBSCAN of Event-based Camera Data with Neuromorphic Computing
\thanks{This material is based upon work supported by  the U.S. Department of Energy, Office of Science, Office of Advanced Scientific Computing Research, under award number DE-SC0022566, and in part on research sponsored by Air Force Research Laboratory under agreement number FA8750-21-1-1018. The U.S. Government is authorized to reproduce and distribute reprints for Governmental purposes, notwithstanding any copyright notation thereon. The views and conclusions contained herein are those of the authors and should not be interpreted as representing the official policies or endorsements, either expressed or implied, of Air Force Research Laboratory or the U.S. Government.}
}

\author{
\IEEEauthorblockN{Charles P. Rizzo}
\IEEEauthorblockA{\textit{Department of EECS} \\
\textit{University of Tennessee}\\
Knoxville, TN, USA \\
crizzo@utk.edu}
\and
\IEEEauthorblockN{Catherine D. Schuman}
\IEEEauthorblockA{\textit{Department of EECS} \\
\textit{University of Tennessee}\\
Knoxville, TN, USA \\
cschuman@utk.edu}
\and
\IEEEauthorblockN{James S. Plank
\IEEEauthorblockA{\textit{Department of EECS} \\
\textit{University of Tennessee}\\
Knoxville, TN, USA \\
jplank@utk.edu}
}}

\maketitle

\begin{abstract}
Spiking neural networks are powerful computational elements that pair well with event-based cameras (EBCs). In this work, we present two spiking neural network architectures that process events from EBCs: one that isolates and filters out events based on their speeds, and another that clusters events based on the DBSCAN algorithm.
\end{abstract}

\begin{IEEEkeywords}
spiking neural networks,
neuromorphic computing,
event-based cameras,
dbscan
\end{IEEEkeywords}

\section{Introduction and Related Works}

In neuromorphic computing, spiking neural networks are leveraged as compute modules that typically perform some sort of control-based task or data classification. Spiking neural networks (SNNs), unlike traditional von-Neumann style architectures, feature both the memory and computation elements in the network together. By using spikes to propagate information throughout the network over time, neuromorphic computing draws direct inspiration from how the biological brain functions. By mimicking the human brain in function, neuromorphic computing aims to realize computational elements that are smaller in size and require less power than both conventional computing architectures and artificial neural networks.

Event-based cameras are novel vision sensors dubbed ``neuromorphic cameras'' because of the way they asynchronously emit events, which is similar to SNNs' spike-based computing architecture. An event-based camera's events are analogous to an SNN's spikes. The camera's pixels asynchronously emit events whenever they individually measure a change in light intensity that is greater than some threshold. The cameras are low-power, have high dynamic range, and don't suffer motion blur like conventional frame-based cameras. Their high temporal resolution ($\mu$s) can make them difficult to work with as they can produce large amounts of data quickly. Fortunately, because spiking neural networks are temporal in nature and feature an event-driven compute architecture, they pair naturally with event-based cameras.

Traditionally, as with artificial neural networks (ANNs), spiking neural networks are trained on an instance of a problem so that over time as the networks continue to train, their performance improves. What complicates spiking neural network training is the nondifferentiability of the spike operation which renders traditional backpropagation useless. In response to this, surrogate gradient based learning algorithms like Slayer~\cite{shrestha2018slayer} have helped bridge the gap between spiking neural networks and backpropagation. There are other training methods, though, like the EONS~\cite{smp:20:eons} and LEAP~\cite{coletti2020library} genetic algorithms that apply genetic mutators to populations of networks over time. Other methods to train SNNs like reservoir computing~\cite{reservoir} or Whetstone~\cite{Whetstone}, an algorithm which converts an ANN to an SNN during its training by slowly converting its activation functions to thresholding activation functions have also been used with success.  These methods facilitate SNN training by circumventing the need for the backpropagation of error throughout the network to train its weights.

A far less popular approach for developing spiking neural networks is to hand-tool or architect them. This is unsurprising, however, as it requires in depth knowledge of a neuroprocessor's architecture in order to fully leverage its neuronal, synaptic, and other network-level features. Manually mapping functions to spiking neural networks is a difficult and time-consuming process. Even still, it is done notably in some works like those by Severa et al.~\cite{7738681}, Plank et al.~\cite{pzs:21:snn}, and Rizzo et al.~\cite{rsp:23:nde} for example. Hand-tooled spiking neural networks are typically small and scalable, precisely defined behavior-wise, and can be leveraged as network components in an ensemble. In this work, we present and precisely define two hand-tooled spiking neural network architectures that perform the following functions on event-based camera data:
\begin{enumerate}
    \item Speed filtration where events can be filtered out for moving too quickly or too slowly
    \item Density-based spatial clustering of events based on the DBSCAN~\cite{ester1996density} algorithm
\end{enumerate}

\section{Method}

\subsection{General Network Specification}

Our spiking neurons implement the simple integrate and fire neuron model where a neuron fires when its potential exceeds its threshold \textit{t}. The synapses of the network are simple featuring only discrete weights \textit{w} and delays \textit{d}. Additionally, the neurons have the option of enabling total leak, where at the end of a neuron's cycle, if there is any charge remaining, it is all leaked away and the neuron's potential reset to its resting potential. Other features common in spiking architectures like spike timing dependent plasticity (STDP)~\cite{STDP} for learning are not necessary for this work. In this way, our network specifications are elegant requiring bare-bones hardware resources for each network. Additionally, a neuron can simultaneously serve as both an input and an output neuron on which input spikes may be applied and firings decoded into decisions.

\subsection{Speed Filter}

The speed filter function is a technique that filters out events based on their ``speed''. An input event \textit{x}'s speed is determined by the amount of events that occur within some radius epsilon ($\epsilon$) of event \textit{x} over the course of two timesteps. If some number of events occur within event \textit{x}'s radius that is greater than the speed threshold \textit{t}, we say event \textit{x} is moving fast. However, if the number of events is less than or equal to the speed threshold \textit{t}, we filter out event \textit{x} as moving too slow. Alternatively, we can identify events that are moving slower than the speed threshold \textit{t} and filter out events moving faster than it. This filtering technique is functionally equivalent to the spiking threshold pooling (or counting) networks presented in~\cite{rsp:23:nde} but without the downsampling component. Even though we are technically thresholding neighboring event density around some event \textit{x}, we are doing it over time and with the understanding that events tend to represent regions or objects that are in motion. This is why we denote this technique as the ``speed filter''.

The ``speed'' filter is influenced by~\cite{nagaraj2023dotie}, in which Nagaraj et al. filter events from a DVS camera by constructing a spiking neuron layer whose dimensionality is the same as the camera. For a camera whose resolution is 640 $\times$ 480, this is over 307,000 spiking neurons. Additionally, they define the network's connectivity by ``connecting'' each neuron to nine pixels: the pixel it represents and the eight other pixels that are a Chebyshev distance of 1 away from it. This means that their speed filtration spiking neural network solution has over 2.7 million synapses. Instead, we present the same functionality as a modular network that can be parallelized and deployed to hardware systems that cannot dedicate such an immense quantity of resources to only event filtration.

\subsubsection{Network Specification}

For the network representation of the speed filter, there are three parameters that contribute to its architecture:
\begin{enumerate}
    \item The speed threshold \textit{t}$_{s}$
    \item The radius epsilon $\epsilon_{s}$ (defined as the Chebyshev distance in this work) 
    \item Whether we filter out events that are moving faster or slower than the speed threshold
\end{enumerate}

First, let us assume we filter out events that are moving slower than the speed threshold \textit{t}$_{s}$. In defining $\epsilon_{s}$ as the Chebyshev distance, there are $(2 * \epsilon_{s} + 1)^{2}$ input neurons labeled~$I_{i,j}$ and a single output neuron~$O$. The input neurons~$I_{i,j}$ have thresholds of 0, and the output neuron~$O$ has a threshold equal to the speed threshold \textit{t}$_{s}$. It is worth noting that distance metrics other than the Chebyshev distance may be used (e.g. Euclidean, Mahalanobis, etc.), and only the number of input neurons is affected by this choice.  The connectivity of the network is defined by one set of synapses:
\begin{enumerate}
    \item $I_{i,j}$ to~$O$ with a weight of 1 and a delay of 0
\end{enumerate}

\begin{figure}[ht]
\centering
\includegraphics[width=0.7\linewidth]{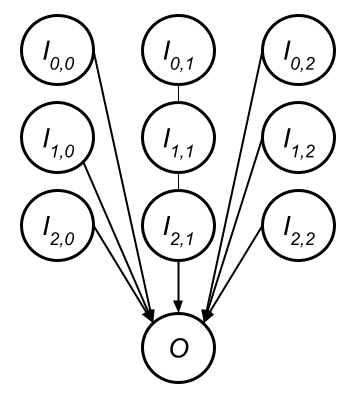}
\caption{Speed filter network architecture with $\epsilon_{s}$ = 1 that filters out events moving slower than speed\_threshold \textit{t}$_{s}$ = 10.}
\label{fig:speed_filter}
\end{figure}

Figure \ref{fig:speed_filter} illustrates what this architecture looks like for~$\epsilon_{s} = 1$ and speed threshold \textit{t}$_{s}$ = 10. 

Let's suppose an event at pixel \textit{x,y} occurs at time \textit{n} and we want to determine whether or not to filter it out based on its speed. The current event's pixel is mapped to input neuron~$I_{\epsilon_{s},\epsilon_{s}}$, and the other adjacent pixels ranging from $x\in [x-\epsilon_{s},x+\epsilon_{s}], y\in [y-\epsilon_{s},y+\epsilon_{s}]$ simply map to the input neurons~$I_{j\in [0,2 * \epsilon] ,i\in [0,2 * \epsilon_{s}]}$. The network cycles required to process the event at pixel $x,y$ are detailed below:
\begin{enumerate}
    \item \textbf{Cycle 0}: Apply events that occur at pixels $x\in [x-\epsilon_{s},x+\epsilon_{s}], y\in [y-\epsilon_{s},y+\epsilon_{s}]$ at time \textit{n - 1} to inputs~$I_{j,i}$.
    \item \textbf{Cycle 1}: Apply events that occur at pixels $x\in [x-\epsilon_{s},x+\epsilon_{s}], y\in [y-\epsilon_{s},y+\epsilon_{s}]$ at time \textit{n} to inputs~$I_{j,i}$. Current event \textit{x,y} is applied to~$I_{\epsilon_{s},\epsilon_{s}}$ during this cycle.
    \item \textbf{Cycle 2}: Output neuron~$O$ accumulates events as charge from events applied at cycle 0. If output neuron~$O$'s potential exceeds its threshold \textit{t}$_{s}$, it fires.
    \item \textbf{Cycle 3}: Output neuron~$O$ accumulates events as charge from events applied at cycle 1. If output neuron~$O$'s potential exceeds its threshold \textit{t}$_{s}$, it fires.
\end{enumerate}

The network requires a total of four cycles to process one event at pixel location $x,y$ after which the network's state may be cleared so that it can process the next event. Table~\ref{tab:speed_filter_ex1} shows an example spike table for Figure~\ref{fig:speed_filter} where the event is not filtered out, and Table~\ref{tab:speed_filter_ex2} shows an example spike table for Figure~\ref{fig:speed_filter} where the event is filtered out.

\begin{table*}[]
\centering
\caption{Example for speed filter shown in Figure~\ref{fig:speed_filter}. Current event of interest is applied to I$_{1,1}$ at timestep 1. Because more than the speed\_threshold \textit{t$_{s}$ = 10} events are applied between timesteps 0 and 1, output \textit{O} fires at timestep 3. The event is not filtered out and passes through the network for downstream processing.}
\label{tab:speed_filter_ex1}
\begin{tabular}{|l|c|cccccccc|}
\hline
\multicolumn{1}{|c|}{Timestep} & Apply spike to                                                    & \textit{}          &                    & \textit{}          & \textit{Fire}      & \textit{}          & \textit{}          & \textit{}          & \textit{}  \\
                               & \multicolumn{1}{l|}{}                                             & \textit{I$_{0,1}$} & \textit{I$_{1,0}$} & \textit{I$_{1,1}$} & \textit{I$_{1,2}$} & \textit{I$_{2,0}$} & \textit{I$_{2,1}$} & \textit{I$_{2,2}$} & \textit{O} \\ \hline
0                              & I$_{0,1}$, I$_{1,0}$, I$_{1,2}$, I$_{2,1}$, I$_{2,2}$             & -                  & -                  & -                  & -                  & -                  & -                  & -                  & \textit{-} \\
1                              & I$_{1,0}$,  I$_{1,1}$, I$_{1,2}$, I$_{2,0}$, I$_{2,1}$, I$_{2,2}$ & *                  & *                  & -                  & *                  & -                  & *                  & *                  & \textit{-} \\
2                              & -                                                                 & -                  & *                  & *                  & *                  & *                  & *                  & *                  & \textit{-} \\
3                              & -                                                                 & -                  & -                  & -                  & -                  & -                  & -                  & -                  & \textit{*} \\ \hline
\end{tabular}
\end{table*}

\begin{table*}[]
\centering
\caption{Example for speed filter shown in Figure~\ref{fig:speed_filter}. Current event of interest is applied to I$_{1,1}$ at timestep 1. Because less than or equal to the speed\_threshold \textit{t$_{s}$ = 10} events are applied between timesteps 0 and 1, output \textit{O} does not fire. The event is filtered out.}
\label{tab:speed_filter_ex2}
\begin{tabular}{|l|c|cccccccc|}
\hline
\multicolumn{1}{|c|}{Timestep} & Apply spike to                                         & \textit{}          &                    & \textit{}          & \textit{Fire}      & \textit{}          & \textit{}          & \textit{}          & \textit{}  \\
                               & \multicolumn{1}{l|}{}                                  & \textit{I$_{0,1}$} & \textit{I$_{1,0}$} & \textit{I$_{1,1}$} & \textit{I$_{1,2}$} & \textit{I$_{2,0}$} & \textit{I$_{2,1}$} & \textit{I$_{2,2}$} & \textit{O} \\ \hline
0                              & I$_{0,1}$, I$_{1,0}$, I$_{1,2}$, I$_{2,1}$             & -                  & -                  & -                  & -                  & -                  & -                  & -                  & \textit{-} \\
1                              & I$_{1,0}$,  I$_{1,1}$, I$_{1,2}$, I$_{2,0}$, I$_{2,1}$ & *                  & *                  & -                  & *                  & -                  & *                  & -                  & \textit{-} \\
2                              & -                                                      & -                  & *                  & *                  & *                  & *                  & *                  & -                  & \textit{-} \\
3                              & -                                                      & -                  & -                  & -                  & -                  & -                  & -                  & -                  & \textit{-} \\ \hline
\end{tabular}
\end{table*}

If however we want to filter out events that are moving faster than the speed threshold \textit{t}$_{s}$, we must slightly tweak the network architecture. In addition to the architecture already described, we add two neurons and three synapses. The first neuron is an additional bias input neuron we denote as~$I_{b}$ whose threshold \textit{t = 0}. The second additional neuron is the new output neuron that we denote as~$O_{n}$ whose threshold \textit{t = 2}. In our previous architecture, neuron~$O$ was our output neuron, but in this architecture, it is a hidden neuron. The connectivity is the same as described above but with the following additions:
\begin{enumerate}
    \item $I_{b}$ to $I_{b}$: This is a self-edge whose synaptic weight is 1 and delay is 0. This ensures that once neuron $I_{b}$ fires, it continues to fire at each cycle until the end of the network simulation.
    \item $I_{b}$ to $O_{n}$ with a weight of 1 and a delay of 0
    \item $O$ to $O_{n}$ with an inhibitory synaptic weight of -1 and delay of 0
\end{enumerate}

\begin{figure}[ht]
\centering
\includegraphics[width=0.8\linewidth]{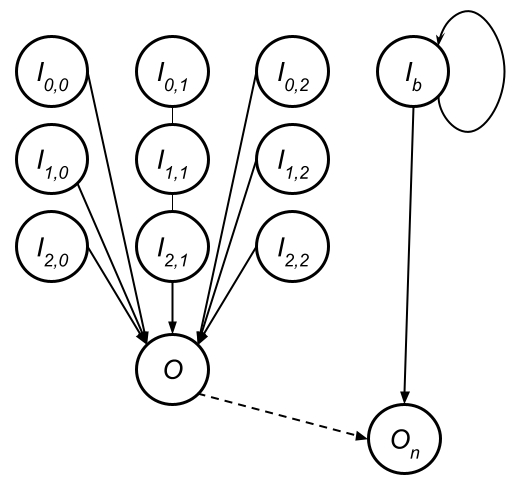}
\caption{Speed filter network architecture with $\epsilon_{s}$ = 1 that filters out events moving faster than speed\_threshold \textit{t}$_{s}$ = 7.}
\label{fig:speed_filter_inverse}
\end{figure}

Figure~\ref{fig:speed_filter_inverse} shows this modified architecture when $\epsilon = 1$ and the speed threshold \textit{t} = 10. The dashed line in Figure~\ref{fig:speed_filter_inverse} and latter figures represents an inhibitory synapse.

Again, let's suppose an event at pixel \textit{x,y} occurs at time \textit{n} and we want to determine whether or not to filter it out based on its speed being too fast. The network cycles required to process the event at pixel $x,y$ are detailed below:
\begin{enumerate}
    \item \textbf{Cycle 0}: Apply events that occur at pixels $x\in [x-\epsilon_{s},x+\epsilon_{s}], y\in [y-\epsilon_{s},y+\epsilon_{s}]$ at time \textit{n - 1} to inputs~$I_{j,i}$. Additionally, we apply a spike to the new bias neuron~$I_{b}$
    \item \textbf{Cycle 1}: Apply events that occur at pixels $x\in [x-\epsilon_{s},x+\epsilon_{s}], y\in [y-\epsilon_{s},y+\epsilon_{s}]$ at time \textit{n} to inputs~$I_{j,i}$. Current event \textit{x,y} is applied to~$I_{\epsilon_{s},\epsilon_{s}}$ during this cycle. Bias input neuron~$I_{b}$ fires sending charge to itself and to output neuron~$O_{n}$.
    \item \textbf{Cycle 2}: Hidden neuron~$O$ accumulates charge from events applied at cycle 0. If hidden neuron~$O$'s potential exceeds its threshold \textit{t}$_{s}$, it fires with an inhibitory spike to output neuron~$O_{n}$. Bias input neuron~$I_{b}$ fires sending charge to itself and to output neuron~$O_{n}$.
    \item \textbf{Cycle 3}: Hidden neuron~$O$ accumulates charge from events applied at cycle 1. If hidden neuron~$O$'s potential exceeds its threshold \textit{t}$_{s}$, it fires with an inhibitory spike to output neuron~$O_{n}$. Bias input neuron~$I_{b}$ fires sending charge to itself and to output neuron~$O_{n}$.
    \item \textbf{Cycle 4}: If no inhibitory spikes arrive at output neuron~$O_{n}$ from hidden neuron~$O$, ~$O_{n}$ fires meaning that we do not filter out the event. If instead one or more inhibitory units of charge arrive at neuron~$O_{n}$ from neuron~$O$, the output neuron will not fire, and the event at \textit{x,y} is filtered out. 
\end{enumerate}

\begin{table*}[]
\centering
\caption{Example for speed filter inverse functionality shown in Figure~\ref{fig:speed_filter_inverse}. Current event of interest is applied to I$_{1,1}$ at timestep 1. A spike is applied to the bias neuron I$_{b}$ at timetstep 0, and it fires into itself and output O$_{n}$ in each successive timestep. Because more than the speed\_threshold \textit{t = 7} events are applied between timesteps 0 and 1, hidden neuron \textit{O} fires at timestep 3. This fire directly inhibits output O$_{n}$ from firing, and the event of interest is filtered out.}
\label{tab:speed_filter_inverse_ex1}
\begin{tabular}{|l|c|cccccccc|}
\hline
\multicolumn{1}{|c|}{Timestep} & Apply spike to                                         & \textit{}        &                    & \textit{}          & \textit{Fire}      & \textit{}          & \textit{}          & \textit{}  & \textit{}        \\
                               & \multicolumn{1}{l|}{}                                  & \textit{I$_{b}$} & \textit{I$_{1,0}$} & \textit{I$_{1,1}$} & \textit{I$_{1,2}$} & \textit{I$_{2,0}$} & \textit{I$_{2,1}$} & \textit{O} & \textit{O$_{n}$} \\ \hline
0                              & I$_{b}$, I$_{1,0}$, I$_{1,2}$, I$_{2,1}$               & -                & -                  & -                  & -                  & -                  & -                  & -          & \textit{-}       \\
1                              & I$_{1,0}$,  I$_{1,1}$, I$_{1,2}$, I$_{2,0}$, I$_{2,1}$ & *                & *                  & -                  & *                  & -                  & *                  & -          & \textit{-}       \\
2                              & -                                                      & *                & *                  & *                  & *                  & *                  & *                  & -          & \textit{-}       \\
3                              & -                                                      & *                & -                  & -                  & -                  & -                  & -                  & *          & \textit{-}       \\
4                              & -                                                      & *                & -                  & -                  & -                  & -                  & -                  & -          & -                \\ \hline
\end{tabular}
\end{table*}

\begin{table*}[]
\centering
\caption{Example for speed filter inverse functionality shown in Figure~\ref{fig:speed_filter_inverse}. Current event of interest is applied to I$_{1,1}$ at time step 1. A spike is applied to the bias neuron I$_{b}$ at timetstep 0, and it fires into itself and output O$_{n}$ in each successive timestep. Because less than or equal to the speed\_threshold \textit{t = 7} events are applied between timesteps 0 and 1, hidden neuron \textit{O} does not fire. Instead, by timestep 4,  I$_{b}$ has contributed enough charge to output O$_{n}$ for it to fire, allowing the event to be passed through and not filtered out.}
\label{tab:speed_filter_inverse_ex2}
\begin{tabular}{|l|c|cccccccc|}
\hline
\multicolumn{1}{|c|}{Timestep} & Apply spike to                           & \textit{}        &                    & \textit{}          & \textit{Fire}      & \textit{}          & \textit{}          & \textit{}  & \textit{}        \\
                               & \multicolumn{1}{l|}{}                    & \textit{I$_{b}$} & \textit{I$_{1,0}$} & \textit{I$_{1,1}$} & \textit{I$_{1,2}$} & \textit{I$_{2,0}$} & \textit{I$_{2,1}$} & \textit{O} & \textit{O$_{n}$} \\ \hline
0                              & I$_{b}$, I$_{1,0}$, I$_{1,2}$, I$_{2,1}$ & -                & -                  & -                  & -                  & -                  & -                  & -          & \textit{-}       \\
1                              & I$_{1,0}$,  I$_{1,1}$, I$_{1,2}$         & *                & *                  & -                  & *                  & -                  & *                  & -          & \textit{-}       \\
2                              & -                                        & *                & *                  & *                  & *                  & -                  & -                  & -          & \textit{-}       \\
3                              & -                                        & *                & -                  & -                  & -                  & -                  & -                  & -          & \textit{-}       \\
4                              & -                                        & *                & -                  & -                  & -                  & -                  & -                  & -          & *                \\ \hline
\end{tabular}
\end{table*}

Filtering out events that are moving faster than the threshold \textit{t}$_{s}$ requires that the network run for one additional cycle per event or five total cycles per event. Table~\ref{tab:speed_filter_inverse_ex1} shows an example spike table for Figure~\ref{fig:speed_filter_inverse} where the event is filtered out, and Table~\ref{tab:speed_filter_inverse_ex2} shows an example spike table for Figure~\ref{fig:speed_filter_inverse} where the event is filtered out. In summary, the following resources are required:
\begin{itemize}
    \item To filter out events moving slower than speed threshold \textit{t}$_{s}$:
    \begin{itemize}
        \item \textbf{Neuron Count}: $(2 * \epsilon_{s} + 1)^{2} + 1$ neurons
        \item \textbf{Synapse Count}: $(2 * \epsilon_{s} + 1)^{2}$ synapses
        \item \textbf{Runtime per Event}: 4 cycles
    \end{itemize}
    \item To filter out events moving faster than speed threshold \textit{t}$_{s}$:
    \begin{itemize}
        \item \textbf{Neuron Count}: $(2 * \epsilon_{s} + 1)^{2} + 3$ neurons
        \item \textbf{Synapse Count}: $(2 * \epsilon_{s} + 1)^{2} + 3$ synapses
        \item \textbf{Runtime per Event}: 5 cycles
    \end{itemize}
\end{itemize}

\subsection{Density-Based Spatial Clustering of Applications with Noise Filter}

Density-based spatial clustering of applications with noise (DBSCAN)~\cite{ester1996density} is a spatial clustering algorithm. It works by clustering all data points as either a core point, a border point, or a noise point. It clusters data based on the premise that spatially close data belongs to the same cluster. The clusters are not rigidly shaped and are composed of core points with border points forming the boundary of the cluster. For our use case, an event is a data point, and it is classified as either core, border, or noise based on the following:
\begin{itemize}
    \item An event is a core event if in its radius epsilon ($\epsilon$), at least threshold \textit{t} (canonically referred to as min\_points) events occur
    \item An event is a border event if in its radius epsilon ($\epsilon$), between one and threshold \textit{t} events occur where at least one of the events is a known core event
    \item An event is a noise event if neither of the prior conditions are met
\end{itemize}

Our network architecture has been validated against scikit-learn's~\cite{pedregosa2011scikit} DBSCAN algorithm across a few different datasets including a few sequences from the DSEC dataset~\cite{Gehrig21ral}.

\subsubsection{Network Specification}
For the network representation of the DBSCAN algorithm, there are only two parameters that contribute to its architecture:
\begin{enumerate}
    \item The threshold \textit{t}$_{d}$ or min\_points value
    \item The radius epsilon $\epsilon_{d}$ defined as the Chebyshev distance 
\end{enumerate}

Optionally, we can choose to recall and label noise events, but we argue that it makes more sense to filter them out as noise. Additionally, for our implementation of the DBSCAN algorithm neuromorphically, the clusters are not labelled and identified. The events are only classified as either core or border events while noise events are filtered out. 

We define two sets of input neurons~$I_{i,j}$ and~$I_{a,b}$. Each set of input neurons, like with the speed filter network, is composed of $(2 * \epsilon_{d} + 1)^{2}$ neurons, which results in a total of $(2 * \epsilon_{d} + 1)^{2} * 2$ input neurons. The input neurons have thresholds \textit{t = 0}. There are four hidden neurons and two output neurons. The output neuron~$O_{c}$ is the core output neuron and its threshold \textit{t} = 0. The border output neuron is denoted as~$O_{b}$ and its threshold \textit{t} = 1. There are also four hidden neurons denoted as~$H_{i \in [0,3]}$ whose thresholds differ and will be explained in a later example that walks through the network activity. The connectivity is defined as follows:

\begin{enumerate}
    \item $I_{i,j}$ to~$H_{0}$ with a weight of 1 and delay of 0
    \item $I_{i,j}$ to~$H_{1}$ with a weight of 1 and delay of 0
    \item $I_{a,b}$ to~$H_{3}$ with a weight of 1 and delay of 0
    \item $H_{0}$ to~$H_{2}$, $H_{1}$ to~$O_{c}$, $H_{2}$ to~$O_{b}$, $O_{c}$ to~$O_{c}$, $H_{3}$ to~$O_{b}$ with a weight of 1 and delay of 0
    \item $H_{1}$ to~$H_{2}$, $O_{c}$ to~$O_{b}$ with a weight of -1 and delay of 0
\end{enumerate}

\begin{figure}[ht]
\centering
\includegraphics[width=1.05\linewidth]{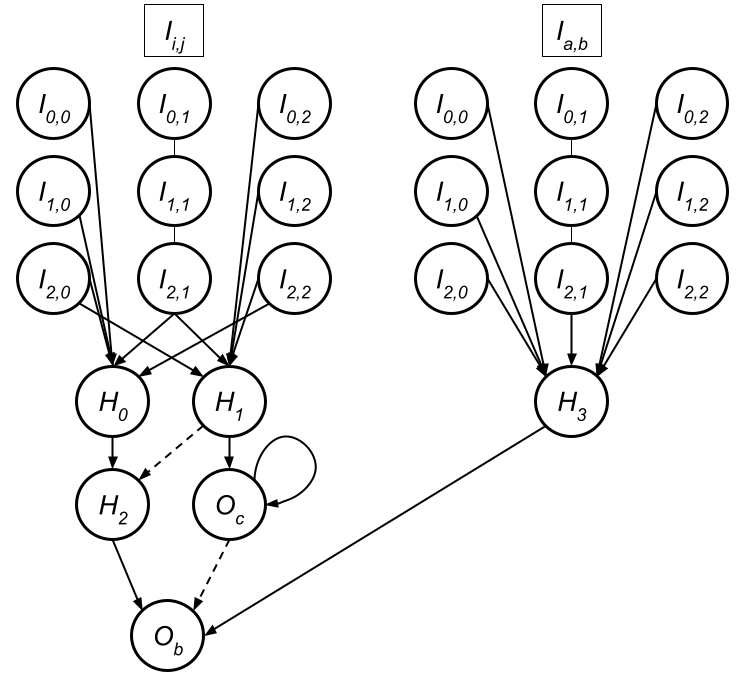}
\caption{DBSCAN network architecture with $\epsilon_{d}$ = 1 and threshold \textit{t}$_{d}$ = 3.}
\label{fig:dbscan}
\end{figure}

To better illustrate the connectivity, Figure~\ref{fig:dbscan} shows a network that performs the DBSCAN algorithm with an $\epsilon_{d}$ value of 1 and a threshold \textit{t} or min\_points value of 3.

Like with our speed filter network example, let's suppose an event occurs at pixel \textit{x,y} and time \textit{n}, and we want to determine whether this event is a core point event, a border point event, or a noise point event. Again, the current event's pixel is mapped to input neuron~$I_{\epsilon_{d} \in j,\epsilon_{d} \in i}$, and the other adjacent pixels ranging from $x\in [x-\epsilon_{d},x+\epsilon_{d}], y\in [y-\epsilon_{d},y+\epsilon_{d}]$ simply map to the input neurons~$I_{j\in [0,2 * \epsilon_{d}] ,i\in [0,2 * \epsilon_{d}]}$. 

Functionally, the inputs~$I_{a,b}$ are equivalent to the input neurons~$I_{i,j}$; however, instead of operating on the event of interest at pixel \textit{x,y},~$I_{a,b}$ operates on each event that occurs in the radius $\epsilon_{d}$ from the event at pixel \textit{x,y}. This is necessary because while inputs~$I_{i,j}$ classify whether the event at \textit{x,y} is a core, border, or noise point, the inputs~$I_{a,b}$ must evaluate whether any of \textit{x,y}'s neighboring events are core point events to validate a border point classification from inputs~$I_{i,j}$ so as to accurately perform the DBSCAN algorithm neuromorphically. Additionally, hidden neuron $H_{3}$ has total leak enabled so that at the end of every cycle, if there is any remaining potential on the neuron, it is leaked away so that we may continue processing each neighboring event each cycle without delay.

The network activity is roughly detailed below:
\begin{enumerate}
    \item \textbf{Cycle 0}: Apply events that occur at pixels $x\in [x-\epsilon_{d},x+\epsilon_{d}], y\in [y-\epsilon_{d},y+\epsilon_{d}]$ at time \textit{n} to inputs~$I_{j,i}$ where the event at pixel \textit{x,y} maps to input I$_{\epsilon_{d} \in j, \epsilon_{d} \in i}$. Additionally, apply the first neighboring event \textit{x',y'} to~$I_{a,b}$ where \textit{x',y'} maps to I$_{\epsilon_{d} \in b, \epsilon_{d} \in a}$; for each subsequent cycle, apply the next event in \textit{x,y}'s neighboring list of events to the~$I_{a,b}$ input neurons.
    \item \textbf{Cycle 2}: 
    \begin{itemize}
        \item If hidden neuron~$H_{1}$ fires, our event at \textit{x,y} is a core event. It in turn fires into output neuron~$O_{c}$ which will continuously generate output spikes and send spikes with inhibitory charge to output neuron~$O_{b}$ to prevent a border point event classification.
        \item If hidden neuron~$H_{1}$ does not fire and only~$H_{0}$ fires, we have a potential border point event classification that must be corroborated by the inputs~$I_{a,b}$ and hidden neuron $H_{3}$ over the course of the next $t_{d} - 1$ cycles.
        \item If neither~$H_{0}$ nor $H_{1}$ fire, the event at pixel \textit{x,y} is classified as noise and filtered out.
    \end{itemize}
    \item \textbf{Cycle 3 to Cycle (min\_points - 1)}: Input neurons~$I_{a,b}$ continue processing events in list of neighboring events $x\in [x-\epsilon_{d},x+\epsilon_{d}], y\in [y-\epsilon_{d},y+\epsilon_{d}]$ (where each event is applied at~$I_{\epsilon_{d} \in b,\epsilon_{d} \in a}$ in the~$I_{a,b}$ input neurons). Supposing one of the events occurs at pixel location \textit{x',y'}, that event is applied at~$I_{\epsilon_{d} \in b,\epsilon_{d} \in a}$ and events that occur at pixels $x\in [x'-\epsilon_{d},x'+\epsilon_{d}], y\in [y'-\epsilon_{d},y'+\epsilon_{d}]$ are mapped to and applied to~$I_{b,a}$ similarly to how the event at pixel \textit{x,y} and its neighboring events in radius $\epsilon_{d}$ are mapped to and applied to~$I_{j,i}$.
\end{enumerate}

\begin{figure}
\centering
\begin{subfigure}{0.35\textwidth}
   \includegraphics[width=1\linewidth]{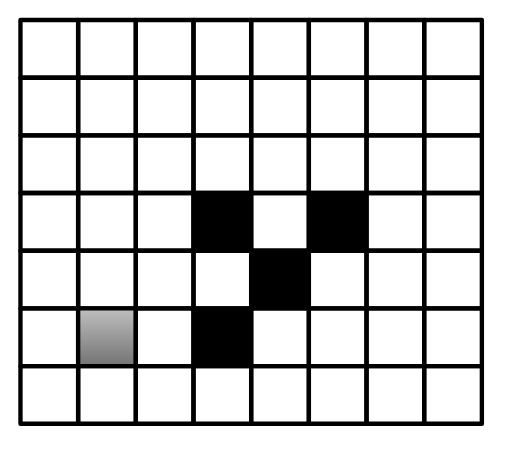}
   \caption{Example for classification of a noise event.}
   \label{fig:noise_dbscan_ex} 
\end{subfigure}

\begin{subfigure}{0.35\textwidth}
   \includegraphics[width=1\linewidth]{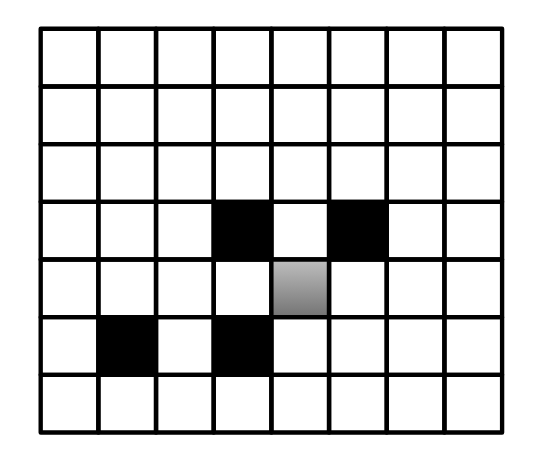}
   \caption{Example for classification of a core event.}
   \label{fig:core_dbscan_ex}
\end{subfigure}

\begin{subfigure}{0.35\textwidth}
   \includegraphics[width=1\linewidth]{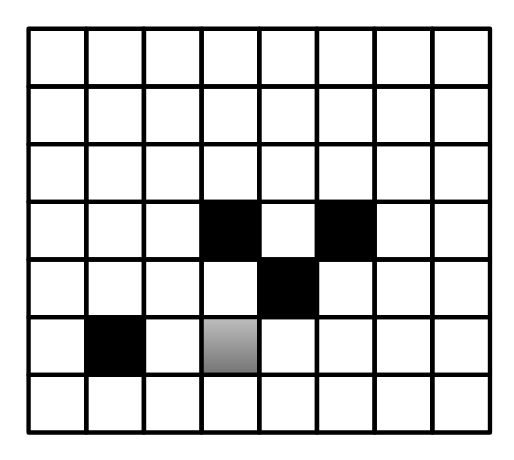}
   \caption{Example for classification of a border event.}
   \label{fig:border_dbscan_ex}
\end{subfigure}
\caption{}
\label{fig:dbscan_examples}
\end{figure}

Three more detailed examples are shown in ~\cref{tab:noise_example_1,tab:core_example_1,tab:core_example_2,tab:border_example_1,tab:border_example_2}. Two tables are needed per example (with the exception of the noise classification), one for the I$_{i,j}$ inputs and one for the I$_{a,b}$ inputs. The examples that they illustrate are shown in Figure~\ref{fig:dbscan_examples}. They show an event's classification as either noise, border, or core based on its adjacent activity.

\begin{table}[]
\centering
\caption{Example spike table for the I$_{i,j}$ input neurons for Figure~\ref{fig:noise_dbscan_ex}. There are no adjacent events, so we omit the table showing the I$_{a,b}$ neurons' activity as there is none.}
\label{tab:noise_example_1}
\begin{tabular}{|l|c|ccc|}
\hline
\multicolumn{1}{|c|}{Timestep} & Apply spike to &  & \textit{Fire} &  \\
\textit{} & \multicolumn{1}{l|}{\textit{}} & \textit{I$_{1,1}$} & \textit{H$_{0}$} & \textit{H$_{1}$} \\ \hline
0 & I$_{1,1}$ & - & - & - \\
1 & - & * & - & - \\
2 & - & - & - & - \\
3 & - & - & - & - \\
4 & - & - & - & - \\
5 & - & - & - & - \\
6 & - & - & - & - \\ \hline
\end{tabular}
\end{table}

\begin{table*}[]
\centering
\caption{Example spike table for the I$_{i,j}$ input neurons for Figure~\ref{fig:core_dbscan_ex}. Since min\_points = 3, and there are a total of 4 events within an epsilon $\epsilon_{d}$ of pixel \textit{x,y} (including itself), both hidden neurons H$_{0}$ and H$_{1}$ fire. H$_{1}$ inhibits neuron H$_{2}$ from firing and causes core output neuron O$_{c}$ to fire. O$_{c}$ fires every cycle afterward because of its self-edge and continuously inhibits output neuron O$_{b}$ from firing, as it still accumulates charge from the I$_{a,b}$ neurons. At the end of simulation, only output O$_{c}$ would have fired, so the classification would be a core neuron classification.}
\label{tab:core_example_1}
\begin{tabular}{|l|c|ccccccccc|}
\hline
\multicolumn{1}{|c|}{Timestep} & Apply spike to & \textit{} & \textit{} & \textit{} & \textit{} & \textit{Fire} & \textit{} & \multicolumn{1}{l}{\textit{}} & \multicolumn{1}{l}{\textit{}} & \multicolumn{1}{l|}{\textit{}} \\
 & \multicolumn{1}{l|}{} & \textit{I$_{0,0}$} & \textit{I$_{0,2}$} & \textit{I$_{1,1}$} & \textit{I$_{2,0}$} & \textit{H${0}$} & \textit{H${1}$} & \multicolumn{1}{l}{\textit{H${2}$}} & \multicolumn{1}{l}{\textit{O${c}$}} & \multicolumn{1}{l|}{\textit{O${b}$}} \\ \hline
0 & I$_{0,0}$, I$_{0,2}$, I$_{1,1}$, I$_{2,0}$ & - & - & - & - & - & - & - & - & - \\
1 & - & * & * & * & * & - & - & - & - & - \\
2 & - & - & - & - & - & * & * & - & - & - \\
3 & - & - & - & - & - & - & - & - & * & - \\
4 & - & - & - & - & - & - & - & - & * & - \\
5 & - & - & - & - & - & - & - & - & * & - \\
6 & - & - & - & - & - & - & - & - & * & - \\ \hline
\end{tabular}
\end{table*}

\begin{table*}[]
\centering
\caption{Example spike table for the I$_{a,b}$ input neurons for Figure~\ref{fig:core_dbscan_ex}. Since $t_{d}$ or min\_points = 3, there's a maximum of 3 events that will need to be processed by the I$_{a,b}$ neurons. We process the events adjacent to \textit{x,y} in row-major order excluding event \textit{x,y}. Because hidden neuron H$_{3}$ has total leak enabled, we can process all adjacent events in succession with no downtime.  For all three adjacent events, there are only two events to apply at the I$_{a,b}$ neurons. This means that no adjacent event will cause neuron H$_{3}$ to fire. }
\label{tab:core_example_2}
\begin{tabular}{|l|c|ccccccc|}
\hline
\multicolumn{1}{|c|}{Timestep} & Apply spike to & \textit{} & \textit{} & \textit{} & \textit{Fire} &  & \multicolumn{1}{l}{\textit{}} & \multicolumn{1}{l|}{\textit{}} \\
 & \multicolumn{1}{l|}{} & \textit{I$_{0,0}$} & \textit{I$_{0,2}$} & \textit{I$_{1,1}$} & \textit{I$_{2,0}$} & \multicolumn{1}{l}{I$_{2,2}$} & \multicolumn{1}{l}{\textit{H${3}$}} & \multicolumn{1}{l|}{\textit{O${b}$}} \\ \hline
0 & I$_{1,1}$, I$_{2,2}$ & - & - & - & - & - & - & - \\
1 & I$_{1,1}$, I$_{2,0}$ & - & - & * & - & * & - & - \\
2 & I$_{1,1}$, I$_{0,2}$ & - & - & * & * & - & - & - \\
3 & - & - & * & * & - & - & - & - \\
4 & - & - & - & - & - & - & - & - \\
5 & - & - & - & - & - & - & - & - \\
6 & - & - & - & - & - & - & - & - \\ \hline
\end{tabular}
\end{table*}

\begin{table*}[]
\centering
\caption{Example spike table for the I$_{i,j}$ input neurons for Figure~\ref{fig:border_dbscan_ex}. We apply the two events to the I$_{i,j}$ neurons which causes hidden neuron H$_{0}$ to fire. Hidden neuron H$_{1}$ does not fire as its threshold has not been exceeded. H$_{2}$ then fires sending charge to O$_{b}$ which fires because it has already received its supplementary required charge from the I$_{a,b}$ neurons at timestep 3 as is shown in Table~\ref{tab:border_example_2}. }
\label{tab:border_example_1}
\begin{tabular}{|l|c|ccccc|}
\hline
\multicolumn{1}{|c|}{Timestep} & Apply spike to & \textit{} & \textit{} & \textit{Fire} & \multicolumn{1}{l}{\textit{}} & \multicolumn{1}{l|}{\textit{}} \\
 & \multicolumn{1}{l|}{} & \textit{I$_{0,2}$} & \textit{I$_{1,1}$} & \multicolumn{1}{l}{H$_{0}$} & \multicolumn{1}{l}{\textit{H$_{2}$}} & \multicolumn{1}{l|}{\textit{O$_{b}$}} \\ \hline
0 & I$_{1,1}$, I$_{0,2}$ & - & - & - & - & - \\
1 & - & * & * & - & - & - \\
2 & - & - & - & * & - & - \\
3 & - & - & - & - & * & - \\
4 & - & - & - & - & - & * \\
5 & - & - & - & - & - & - \\
6 & - & - & - & - & - & - \\ \hline
\end{tabular}
\end{table*}

\begin{table*}[]
\centering
\caption{Spike table for the I$_{a,b}$ input neurons for Figure~\ref{fig:border_dbscan_ex}. We process the sole neighboring event, and apply its three adjacent events to the I$_{a,b}$ neurons which causes hidden neuron H$_{3}$ to fire. Hidden neuron H$_{3}$ sends priming charge to output neuron O$_{b}$ which doesn't fire until cycle 4, which is when it receives its supplemental charge from the I$_{i,j}$ neurons as is shown in Table~\ref{tab:border_example_1}. }
\label{tab:border_example_2}
\begin{tabular}{|l|c|cccccc|}
\hline
\multicolumn{1}{|c|}{Timestep} & Apply spike to & \textit{} & \multicolumn{1}{l}{} & \textit{Fire} & \textit{} & \multicolumn{1}{l}{\textit{}} & \multicolumn{1}{l|}{\textit{}} \\
 & \multicolumn{1}{l|}{} & \textit{I$_{0,0}$} & \multicolumn{1}{l}{\textit{I$_{0,2}$}} & \textit{I$_{1,1}$} & \multicolumn{1}{l}{I$_{2,0}$} & \multicolumn{1}{l}{\textit{H$_{3}$}} & \multicolumn{1}{l|}{\textit{O$_{b}$}} \\ \hline
0 & I$_{0,0}$,I$_{0,2}$I$_{1,1}$, I$_{2,0}$ & - & - & - & - & - & - \\
1 & - & * & * & * & * & - & - \\
2 & - & - & - & - & - & * & - \\
3 & - & - & - & - & - & - & - \\
4 & - & - & - & - & - & - & * \\
5 & - & - & - & - & - & - & - \\
6 & - & - & - & - & - & - & - \\ \hline
\end{tabular}
\end{table*}

This process of classifying events as either core point events, border point events, or noise point events requires threshold (or min\_points) $t + 4$ cycles per event. Classifying an event as noise can be accomplished in 3 cycles, classifying an event as a core point can be done in 4 cycles, but classifying an event as a border point event can require up to threshold (or min\_points) $t_{d} + 4$ cycles. This is due to the fact that classifying an event as a border point event requires the parallel classification of at least threshold (or min\_points) $t_{d} - 1$  neighboring events. The intuition behind this is that if we believe an event at \textit{x,y} is a border point event, there are threshold (or min\_points) $t_{d} - 1$ or less events in its radius $\epsilon_{d}$. We must then process in parallel up to a possible threshold (or min\_points) $t_{d} - 1$ events at the~$I_{a,b}$ neurons to determine if any of them are a core event point. In summary, the following resources are required:
\begin{itemize}
    \item \textbf{Neuron Count}: $(2 * \epsilon_{d} + 1)^{2} * 2 + 6$ neurons
    \item \textbf{Synapse Count}: $(2 * \epsilon_{d} + 1)^{2} * 3 + 7$ synapses
    \item \textbf{Runtime per Event}: threshold (or min\_points) \textit{t}$_{d}$ + 4  cycles
\end{itemize}

\section{Discussion}

When applying spiking neural networks to neuromorphic hardware, there are several practicalities that should be taken into consideration. One of these is the encoding function used to apply the data to the inputs of the neural network. Fortunately, for event-based camera data, this operation is fairly straightforward. An event directly translates to a one-valued spike applied to the appropriate input neuron, regardless of its polarity. Additionally, after processing an event, the network's leftover state needs to be cleared before processing the next event. This soft reset requires a few cycles to reinitialize a network's state before processing the next event. Decoding the output neurons' spikes also must be addressed. For both presented networks, the popular Winner-take-all (WTA) decoding scheme results in the desired network functionalities. In the WTA decoding scheme, whichever output neuron fires the most is the ``winner'', and its classification or action is chosen.

Applying one of the filtering networks to a dataset is straightforward. However, when two or more filtering networks are applied successively, it is more complicated. One option is to time multiplex the observation space and route its inputs through one network, collect that network's outputs off of the neuromorphic chip, and then route the first network's outputs as the second or latter network's inputs. This approach works, but is burdened by data movement costs on and off chip. Instead, we propose compiling the networks together. 

As an example, lets assume we would like to pass all of the events through a speed filter and then a subsequent DBSCAN filter to isolate objects moving at a certain speed, toss out any noise and cluster the events. Because we need to pass every event through a speed filter network before applying it to the DBSCAN filter network, the simplest implementation would require a speed filter network for each input of the DBSCAN filter network. In trading space for time, the compiled network would be composed of the DBSCAN network preceded by $(2 * \epsilon_{d} + 1)^{2} * 2$ speed filter networks. In theory, assuming we are filtering out events that are moving slower than the speed filter threshold $t_{s}$, the runtime of the compiled network would be the sum of both network components give or take a few cycles. Additionally, it's possible to avoid duplication of the speed filter network for every DBSCAN filter network input neuron in favor of trading time for space. Instead, one speed filter network each for the I$_{i,j}$ and I$_{a,b}$ neurons can be used as long as the input events and inter-network activity are properly time multiplexed through clever use of synaptic delays. We leave the detailed characterization and verification of these compiled network approaches to a latter work.

\begin{figure}
\centering
\begin{subfigure}{0.45\textwidth}
   \includegraphics[width=1\linewidth]{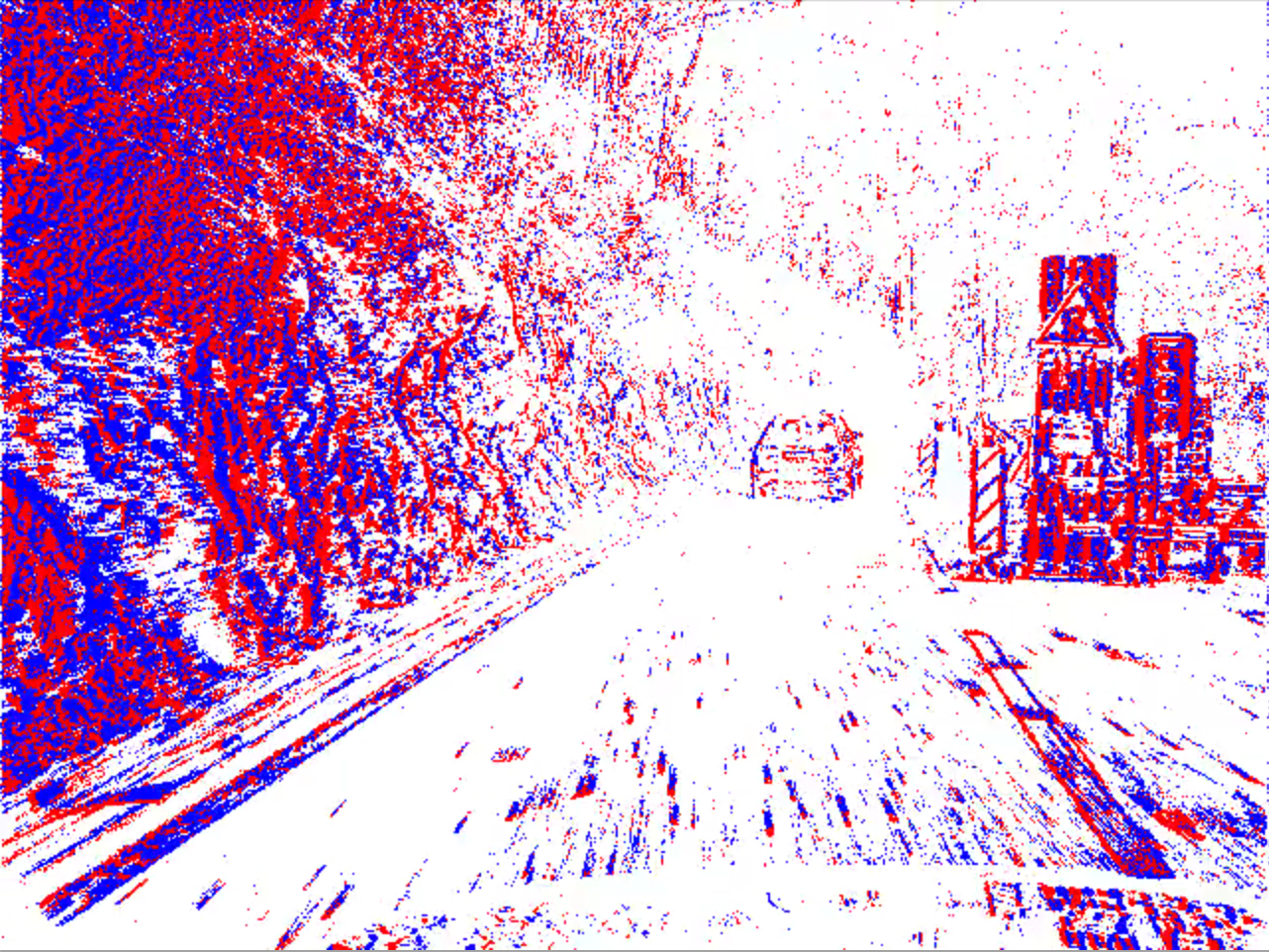}
   \caption{50 ms aggregation of events from the ``interlaken\_00\_c'' sequence.}
   \label{fig:interlaken_before}
\end{subfigure}

\begin{subfigure}{0.45\textwidth}
   \includegraphics[width=1\linewidth]{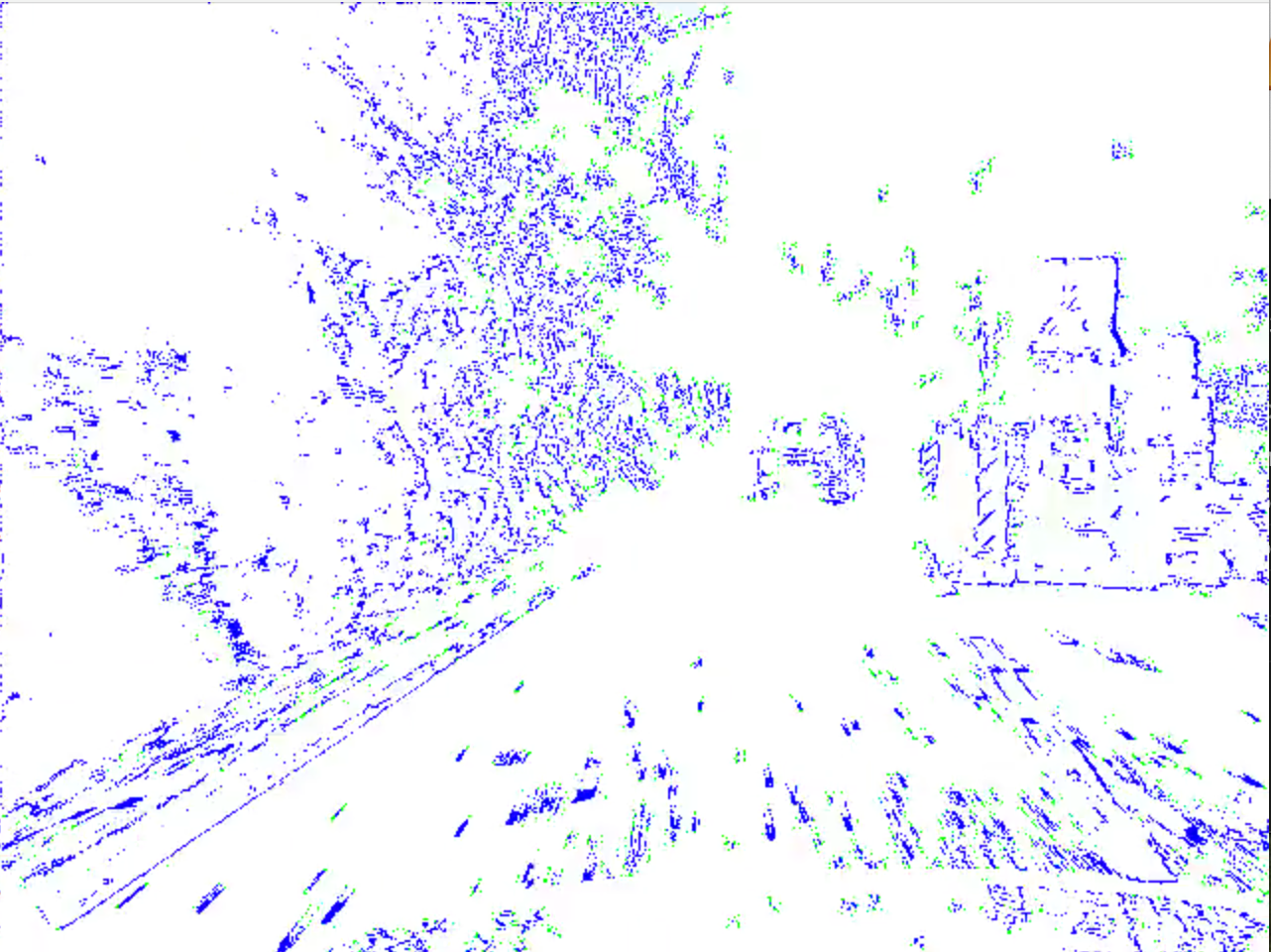}
   \caption{Same as Figure~\ref{fig:interlaken_before} but after the events have been passed through a speed filter and then a DBSCAN filter. The blue events are core events, and the green events are border events.}
   \label{fig:interlaken_after}
\end{subfigure}
\caption{}
\label{fig:interlaken_ex}
\end{figure}

Figure~\ref{fig:interlaken_ex} shows the output of a 50ms aggregation of events from the ``interlaken\_00\_c'' sequence from the DSEC~\cite{Gehrig21ral} dataset both before (Figure~\ref{fig:interlaken_before}) and after (Figure~\ref{fig:interlaken_after}) a speed filter and DBSCAN filter are applied to all of the events in the aggregation. This figure illustrates a use case of both the speed filter network and DBSCAN filter network together by showing the reduction in overall event noise and the clustering of events. The parameters used to generate this output for the speed filter network are t$_{s} = 9$ and $\epsilon_{s} = 1$ using the speed filter inverse functionality (i.e. filter out events that are moving too fast). The parameters used for the DBSCAN network are t$_{d} = 10$ and $\epsilon_{d} = 3$.

\section{Conclusions}

Overall, these architectures are scalable and can be adjusted as desired for target hardware. The speed filter network can be used to isolate or filter out events moving at certain speeds with respect to the camera, and the DBSCAN filter network can be used to spatially cluster events while getting rid of noise. Furthermore, these networks can be used together serially as a system of networks that performs a complex functionality. The end result is a highly scalable, customizable, and parallelizable neuromorphic system capable of event denoising and clustering.

\bibliographystyle{IEEEtran}
\bibliography{references}

\end{document}